\documentclass[
conference
]{IEEEtran}

\usepackage{cite}
\usepackage{amsmath,amssymb,amsfonts}
\usepackage{graphicx}
\usepackage{textcomp}
\usepackage{xcolor}
\def\BibTeX{{\rm B\kern-.05em{\sc i\kern-.025em b}\kern-.08em
    T\kern-.1667em\lower.7ex\hbox{E}\kern-.125emX}}
    
\usepackage{float}
\usepackage{hhline}
\usepackage{todonotes}
\usepackage{algorithm}
\usepackage[noend]{algpseudocode}

\begin{document}

\title{GCN-WP -- Semi-Supervised Graph Convolutional Networks for Win Prediction in Esports}

\author{
\IEEEauthorblockN{
Alexander J. Bisberg
}
\IEEEauthorblockA{
\textit{Viterbi School of Engineering, USC} \\
\textit{Information Sciences Institute, USC}\\
Los Angeles, CA USA \\
bisberg@usc.edu}
\and
\IEEEauthorblockN{
Emilio Ferrara
}
\IEEEauthorblockA{
\textit{Viterbi School of Engineering, USC
} \\
\textit{Annenberg School for Communication and Journalism, USC} \\
\textit{Information Sciences Institute, USC}\\
Los Angeles, CA USA \\
emiliofe@usc.edu}
}

\makeatletter
\def\footnoterule{\kern-3\p@
  \hrule \@width 2in \kern 2.6\p@} 
\makeatother
\newcommand{\copyrightnotice}[1]{{%
  \renewcommand{\thefootnote}{}
  \footnotetext[0]{#1}%
}}

\maketitle
\copyrightnotice{\copyright2022 IEEE. Personal use of this material is permitted. Permission from IEEE must be obtained for all other uses, in any current or future media, including reprinting/republishing this material for advertising or promotional purposes, creating new collective works, for resale or redistribution to servers or lists, or reuse of any copyrighted component of this work in other works.}

\begin{abstract}
Win prediction is crucial to understanding skill modeling, teamwork and matchmaking in esports. 
In this paper we propose GCN-WP, a semi-supervised win prediction model for esports based on graph convolutional networks.
This model learns the structure of an esports league over the course of a season (1 year) and makes predictions on another similar league.
This model integrates over 30 features about the match and players and employs graph convolution to classify games based on their neighborhood.
Our model achieves state-of-the-art prediction accuracy when compared to machine learning or skill rating models for LoL.
The framework is generalizable so it can easily be extended to other multiplayer online games.
\end{abstract}


\section{Introduction}

The first skill based matchmaking algorithm was invented in the 1950s, and eponymous named by, Arpad Elo. 
His intention was to produce a ranking for chess players at tournaments and subsequently predict who would win \cite{elo}.
This system uses a single number to represent a player's or team's skill. 
This algorithm involves two main steps, the score expectation estimation followed by the score update. 
The score expectation is \emph{defined} as the win probability of a given team.
Therefore, skill rating and win prediction have been intimately tied since their origins.

The Elo algorithm was designed to be updated after every game or match. 
The expectation of a player's (or team's) skill changes based on the result of each game. 
Although originally intended for an individual playing chess, many have adapted the Elo algorithm to other sports and games of full teams of players \cite{DeLong2011, scope, boice}.
Some of these models use a single score to represent a team's skill, while others attempt to model each individual player \cite{ts, ts2}.

Win prediction can be used to rank individuals or teams in seeding for a tournament. 
In many professional esport and sport leagues, every team has the opportunity to play every other team in the league, usually multiple times throughout the course of a season.
Is there a way we can exploit this graphical structure to improve our understanding of what it takes to win in an esports league?

\begin{figure}[t]
    \centering
    \includegraphics[width=0.95\linewidth]{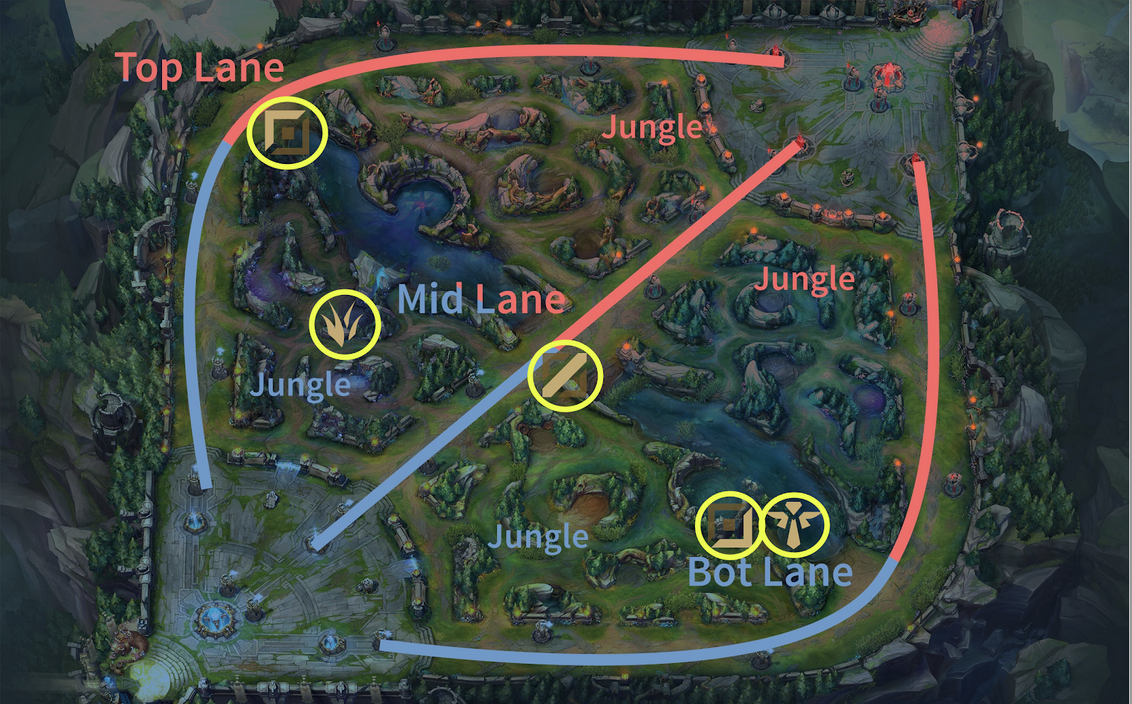}
    \caption{Summoner's Rift - the stage where each game of League of Legends occurs.}
    \label{fig:sr}
\end{figure}

Many rating systems make the assumption that an esports team's performance is connected to their historical results \cite{elo, ts}.
Instead of explicitly modeling their skill over time as current skill models do, we create a graph representation of the entire network of teams to learn how their past games and opponents affect their performance.
Then, we use the order of games and connections between teams, along with features about those games, to predict which team will win. 
There has been promise in analyzing professional soccer league rankings using a networks \cite{football-network}, which enabled comparisons between the various global professional league rankings. 
Graphs have proved useful for analyzing inter-team dynamics in the subject game -- \emph{League of Legends} (LoL) -- \cite{cantallops2019}, but connections between games have yet to be explored. 
Some win prediction models use traditional statistical models to build this graph and make predictions \cite{ts}, but we hypothesized a deep embedding of this graph could yield better win prediction results. 
There has been success using deep learning for win prediction \cite{ghost-recon}, however none have combined graphs with deep learning. 

Given our analysis of the win prediction landscape, we propose using graph neural networks (GNNs) \cite{Scarselli2009} to analyze these professional league graphs. 
GNNs were invented to analyze data poorly suited for Euclidean space. 
In the same way traditional deep models learn feature emgeddings, GNNs learn embeddings of the input graph or graphs. 
There are different techniques to learn these embeddings, one of them involves graph convolutions \cite{Kipf2017}, which we will focus on here. 
Given the context above, we have distilled our aims into the following research questions:

\begin{itemize}
    \item RQ1: Can GNNs improve the performance of win prediction models for elite teams? 
    \item RQ2: Can a league embedding be trained and perform well in a win prediction task on a separate league?
\end{itemize}

To answer these questions we propose GCN-WP: Semi-Supervised \textbf{G}raph \textbf{C}onvolutional \textbf{N}etworks for \textbf{W}in \textbf{P}rediction. This method takes in to account a full seasons worth of historical data so the modeler will not be forced to manually weight past results. 
In addition, graph convolutions are an efficient and elegant way to observe the effect of "nearby" games. 
The key importance of the semi-supervised approach is that the model can learn from data of the current game we are predicting without knowing the outcome of that game. 
This is also the first algorithm that allows for prediction across leagues of esports. 
Without any information about how the teams in the new league will perform, this algorithm shows high prediction accuracy.

\noindent \underline{Contributions} \\
\begin{enumerate}
    \item We demonstrate the capacity for a GCN model to represent a \emph{league} of esports teams, particularly LoL
    and it's ability to capture historical results of those teams and their contribution to a team's ability to win in the future.
    \item This model is compared to other state-of-the-art models in the field and shown to perform better for win prediction at this task. 
    This implies that we can learn from the inherent structure of success in a professional esports league to make predictions about similar leagues. 
\end{enumerate}


\section{Background and Related work}
\label{bkrd}

\subsection{Win Prediction in Esports}

Rating systems are key to understanding the landscape of win prediction models. 
These systems are generally developed to rank players, but also have a component to predict match outcomes. 
Rating systems are useful because they enable statisticians, tournament directors, and fans to quantify player skill. 
In general, a highly skilled individual or player should win against another player that has lower skill. 
Therefore, most skill rating systems have a direct way to predict which individual or team will win. 
To explore the background of \emph{win prediction algorithms} is to explore the background of \emph{rating systems}. 

\subsubsection{Origins: Elo and Gliko}

Arpad Elo was an early pioneer in rating systems. 
The idea was to have a single number represent a player's rating and an equation that could be used to update that rating after a match completed (preferably by hand since calculators were not prolific in the 1950s). 
Thus was born the Elo algorithm, as described in his book \cite{elo}.

Many of the parameters of the Elo algorithm were set by eye without rigorous statistical backing such as cross-validation or back-testing.
He likely didn't have access to enough data to perform these analyses; however,  the United States Chess Federation (USCF) continued to use this algorithm. 
In the 90's a statistician from Harvard, Mark Glickman, questioned the assumptions of Elo's original algorithm. 
He showed that one could add in factors to dyamically adjust the ratings of a population to better fit match outcomes, as well as add a time decay factors to players who haven't played recently. 
Glickman presented most of his findings in his guide on chess ratings \cite{glickman-guide} and improved on win prediction accuracy of Elo's original algorithm for Chess ratings.

\subsubsection{Integrating Players and Teams}

As the growth of online competitive games exploded in the late 2000s, the need for scalable win prediction and ratings systems grew with them. 
Many game companies began developing their own versions of these ratings systems behind closed doors; however, the teams at Microsoft Research and Halo 2 collaborated to release a new rating system called TrueSkill \cite{ts}.
Pioneered by Christopher Bishop and Thomas Minka, two pioneers in probabilistic inference, used Bayes nets to infer player skill as a complex distribution. 
They dramatically increased the complexity of the model wile allowing the developer to add in more modern assumptions. 
TrueSkill models not only each individual's skill over time, but it's \emph{variance} as well. 
To estimate team skill, multiple individual skills are summed or averaged to together.
As the Halo games continued to develop, so did TrueSkill.
The creators released an update aptly named TrueSkill2 \cite{ts2}. 
This version of the algorithm introduced the capacity for automated parameter tuning, integrating other data from players, and modifications based on players who queue up as a squad. 

After TrueSkill came TeamSkill \cite{DeLong2011}, which explicitly attempts to model the skill of a team. 
They achieve this by partition each team of $n$ players in to $k$ subgroups ranging from $0$ to $n$. 
This process is applied to existing Elo, Gliko and Trueskill skill ratings and shows some improvements on model accuracy.

Other research has shown that there are complex and subtle differences when playing esports with a team of players. 
For example, playing with friends may actually benefit low skill players more than high skill players \cite{Zeng2019} and their performance is affected by the number of games they play in a single session as well \cite{Sapienza2018}.
However, these relationships may not hold true for professional esports teams.

In addition, deep learning approaches have been used recently to quantify skill in teams.
NeuralAC and OptMatch \cite{gu2021neuralac, gong2020optmatch} strive to quantify not only interactions between enemy teams, but interactions between allies and character abilities as well. 
\cite{deng2021globally} uses deep reinforcement learning to create a globally optimal matchmaking strategy. 

\subsubsection{Applications to Esports}

As professional esports organizations became more normal across esports from LoL to Call of Duty (CoD) in the mid to late 2010s, the ratings landscaped changed again. 
No longer were ad-hoc teams and tournaments the primary medium for esports to be played.
Leagues much closer to the English Premier League or Major League Baseball began to form. 
The distinction between professional esports ranking and matchmaking is important, and may change the type of skill model one chooses to use \cite{scope}.

Defense of the Ancients 2 (DOTA2) is a game similar to LoL in many aspects, so we investigated which models have been used for this game in particular.
Much like LoL, there are 5 characters (heros, as they are called in DOTA2) on each team battling to destroy each other's base -- more about game mechanics will be explained below in Section \ref{sec:gameplay}.
\cite{semenov2016performance, wang2017outcome, wang2018outcome} all use variations on traditional skill-based models for DOTA2 win prediction or attempt to integrate novel factors like hero draft. 
Some other works do use deep learning for win prediction in LoL (\cite{kim2017makes, kim2020predicting,hitar2022machine}); however, they do not make predictions on professional esports matches.
Finally, \cite{hodge2019win} perform a wide literature review on the win prediction in esports domain showing applications of graph based approaches for integrating the characters players are using in these games. 

\subsection{Neural Networks}

Deep learning is also a promising technique used for win prediction. 
\cite{ghost-recon} show that player embeddings are successful at accurately predicting matches.
Moreover, an advantage of this framework is they consider optimizing for other factors such as player ping or retention. 
Once again, this algorithm is more focused on ad-hoc teams. 

\subsubsection{Graph Neural Networks}

To the best of our knowledge, the GNN model has not been used for esports win prediction tasks before this.
In it's a simplest form, a GNN starts with a graph: nodes (data points) connected by edges \cite{Scarselli2009}. 
A GNN can classify individual nodes in a graph, thereby each node will ahve a label, whose labels will be learned \emph{inductively} from the surrounding nodes. 
Alternatively, GNNs can be trained on multiple labeled graphs to classify an entire graph as opposed to just nodes in the graph.

\subsubsection{Graph Convolution}


\cite{Kipf2017} proposed the graph convolutional network as a way to embed graph data and learn from it. 
Semi-supervised networks are useful when some data are missing labels.
This is similar to the win prediction case when we have data for both of the team's previous games, however we don't have a label for the game we are predicting.

\begin{figure}[h]
    \centering
    \includegraphics[width=0.95\linewidth]{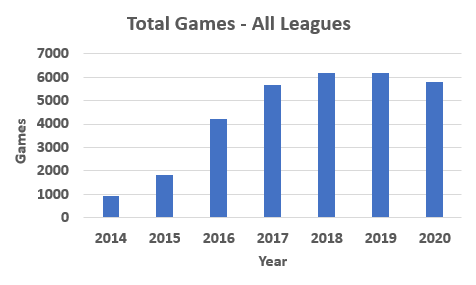}
    \includegraphics[width=0.95\linewidth]{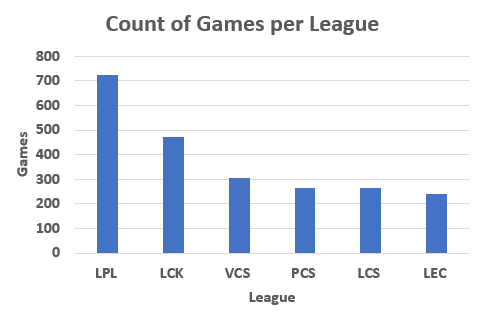}
    \caption{Number of games per year and per league \cite{oe}.}
    \label{fig:games-per}
\end{figure}

\section{Dataset}

\subsection{League of Legends}

\subsubsection{Basic Game Structure} 
\label{sec:gameplay}
As mentioned earlier, LoL is part of the sub-genre Multiplayer Online Battle Arena (MOBA). 
In this case, the players are divided into two teams of five players, and in each of these teams the players select roles and champions. 
The goal of the games is to destroy the enemies' outlying structures called towers, and eventually their base -- the nexus.
Each player controls one avatar (champion) throughout the course of a game -- there are around 170 unique champions to choose from.
Each game of LoL is played on Summoner's Rift \ref{fig:sr}.
This map has 3 lanes, the top lane arcs around the top of the map, the middle bisects the map from corner to corner, and the bottom lane curves around the bottom and side of the map. 
Although the point of the game is to destroy your opponent's base, most of the strategy revolves around gold. 
Gold is earned not only from killing your opponents structures, but also killing enemy champions and creeps. 
Creeps are non-player characters that periodically and symmetrically walk down each lane.
Finally, the jungle is the intermediary area with neutral monsters, shrouded by \emph{fog of war}. 
Players may only see where their allies are on the map, so the junglers move around mostly hidden trying to influence the states of top, middle and bottom lane. 
Eventually in the later stages of the game teams being to group together and fight less in the lanes but more in the jungle around baron and dragon. 
These large monsters take longer to defeat but offer the players gold and buffs to increase their champion strength and allow players to end the game sooner.

\subsubsection{Professional Play}
In 2020 there were six "major" regions in LoL, but this could change from year to year as the strength of a region changes 
In general these major region leagues have the most games played over a season, so  so they are the focus of the following analysis.
As mentioned above, this is a critical component to the proposed model. 
Each league takes place in a different region of the world, however the base game version and rules are the same. 
There may be regional differences in strategies; for example,  in one region kills may be more important, whereas in another towers are prioritized. 
If there are enough similarities between leagues, our model should be able to predict their game outcomes well.
In addition leagues play over seasons, usually corresponding to one year (i.e. the 2018 season).
In each season, a group of teams typically plays other teams from their region.
There are typically a mid-season invitational tournament where the best teams from each league compete in addition to a world championship at the end of the season.
Because these tournaments involve mixing leagues with teams that don't often play each other, we excluded these tournaments from the dataset.

\subsection{Data Collection}
 
This dataset was aggregated by a reputable blog know as Oracle's Elixir managed by  Tim Sevenhuysen, a former esports analyst \cite{oe}.
During each season, some teams are invited to international tournaments, usually by how well they do in the respective regions. 
At the end of the year, the world championships are held.
For this research, we are restricting our analysis to regular season games since during that time, teams are mostly playing other teams from their respective region. 
Therefore, we have a larger sample size of a smaller cohort of teams. 
In addition, this sets up well for a graph neural network framing.

The data has been collected since 2014 to present. 
Figure \ref{fig:games-per} shows a breakdown of the number of games player per year and per league.
For each game played in the professional leagues, this dataset contains much more detailed information than used in traditional win prediction models described in Section \ref{bkrd}. 
The features can be broken up in to categories including \emph{objectives, farm, gold and experience, fighting and vision}. 

\subsection{Features}


\subsubsection{Objectives} are critical aspects of LoL. 
They include towers, inhibitors, and neutral monsters like dragon and baron. 
Destroying towers and inhibitors opens up sections of the map and empowers individual champions to roam more freely, eventually closing in on their opponents base to deal the winning blow.

\subsubsection{Farm} The term farming refers to a team's ability to destroy small neutral monsters on the map, whether they are in the main lanes of the battlefield (creeps) or in the jungle (see Fig \ref{fig:sr} for reference).
Destroying neutral monsters gives teams more gold.

\subsubsection{Gold and Experience} 
Gold is earned for kills, objectives and farm.
With this currency players can buy items to increase their ability to fight or capture objectives.
Experience is earned separately from gold, but also from fighting opponents, neutral monsters, and destroying objectives.

\subsubsection{Fighting} is another critical aspect of the game. 
By slaying an opposing champion, they are prohibited from re-spawning for a certain amount of time that scales with the length of the game. 
Slaying opponents rapidly in quick succession counts as a multi-kill, (double, triple, etc.)
These features could be indicators of a team's fighting prowess. 

\subsubsection{Vision} is a critical component of LoL, especially in pro play. 
Unlike chess, LoL is an \emph{imperfect} information game. 
The battlefield is shrouded in a \emph{fog of war}, where each team can only see where their own units or \emph{wards} are. 
Wards are static and illuminate the section of the map they are placed in. 
This allows for control of neutral objectives in the fog of war like dragon or baron.

We constructed two datasets with these features, one with raw feature values for each team (called \emph{raw} below) and a second with the difference of the feature values between the two teams (called \emph{delta}).

\section{GCN Model Architecture}

\subsection{Building the Graph}

To formulate a GCN for the win prediction task, we must first conceptualize the relationship between sequential games.
We assume that some teams $A$ and $B$ are connected directly to the opponent of their \emph{current} game, and themselves from their previous game.
We propose the algorithm to draw the team graph in Algorithm \ref{alg:graph}. 
For each team the games are ordered in ascending order and a node is added with the features from that game. 
A connection is drawn from that node to an opponents node and to that teams previous node. 
Here we assume a homogeneous network: there is no distinction between self and opponent nodes. 
We believe this is a fair assumption that can be justified with feature engineering by using the \emph{delta} features.
This allows the data representation at each node to be a difference between the team and their opponent, effectively homogenizing the network. 
In addition, the graph convolutions will allow us to learn the structure of the network regardless of the node type.

\begin{algorithm}[tb]
\begin{algorithmic}[1]
\caption{BuildLeagueGraph}
\label{alg:graph}
\State $G \gets (V,E)$ \Comment{initialize graph}
\For{$t$ in $teams$}{:}
\State $games[t] \gets \text{sorted}(games[t])$ \Comment{sorted by time}
\For{$g,i$ in $games$}{:}
\State $G(N) \gets g^t_i$ \Comment{add node for team game}
\State $G(E) \gets (g^t_i, g^{opp}_i)$ \Comment{add edge to opponent}
\State $G(E) \gets (g^t_i, g^{t}_{i-1})$ \Comment{add edge to previous game}
\EndFor
\EndFor
\State \Return $G$
\end{algorithmic}
\end{algorithm}

\subsection{Exploiting Graph Structure}

\begin{figure}[b] 
    \centering
    \includegraphics[width=0.95\linewidth]{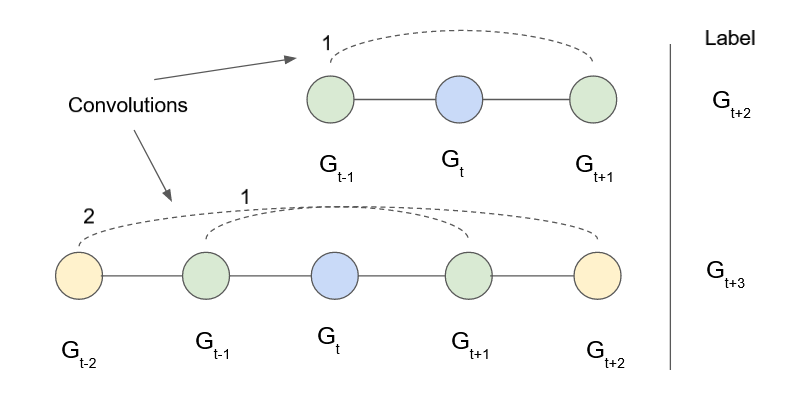}
    \caption{This is an example of how graph convolution affects the labeling of our features. Given the architecture is symmetrical, convolutions will look at future nodes and past nodes. The label for the node at time $t$ is the result of $G_{c+1}$ (win = 1, loss = 0) where $c$ is the number of convolutions. This ensures that data is not leaking out of the model in to the other predictions}
    \label{fig:gconv}
\end{figure}

The main difference in this model from those mentioned in Section \ref{bkrd}, is that we are observing a yearly, scheduled, professional esports leagues as opposed to ad-hoc matches.
The most similar analog to professional sports is soccer. 
Many countries have their own soccer leagues, like Major League Soccer (MLS) in the United States, and the Premier League in Britain. 
Each of these leagues generally play many other teams in the league (if not all of them) in a given season. 
Therefore, we have the most examples of these teams playing \emph{each other}. 
There are also global competitions that pit these teams from different regions against each other, but these happen much more infrequently. 
Our idea is that we can exploit the structure and history of a team in any given league to learn from that particular league, then apply that model to other leagues. 
Therefore, we are effectively training on data different from the final network.
As far as we know this is the first esports win prediction task to attempt this type of prediction.

\subsection{Model Assumptions}

One assumption that many win prediction models make is that more recent performance should be weighted more heavily than earlier performance. 
This is inherent in the construction of SCOPE and TrueSkill, each making sequential updates to a score. 
In a graphical representation, there would be connections from a team's past game nodes to a team's present game node.
More games are observed as the season progresses.
However, at what point do a team's previous results \emph{negatively} impact predictions? 
This is an open question and likely varies for each domain (esport, league, etc.). 
To represent this data in a graph, there is a list of $N$ nodes and their features $f$, along with the relationships between these nodes, typically in the form of an adjacency matrix, $A$.
\cite{Kipf2017} use \emph{convolutions}, or filters, to capture the influence of surrounding nodes on any given node. 
Each sequential neural network layer can be written as a non-linear function \cite{kipf-gcn-blog}:
\[ L^{(n+1)} = f(L^{(n)},A)  \] 
Where $L$ is a layer, $n$ is the layer number and $A$ is the node adjacency matrix.
The idea is that each layer should relate to the original adjacency matrix in some way. More formally, each layer $L$ should be scaled by the adjacency matrix $A$ and the learned weights $W$ then processed through some non-linear activation function $\sigma$ like ReLU (the Rectified Linear Unit, a common neural network activation function).

\[f(L^{(n)},A) = \sigma \left( AL^{(n)}W^{(n)} \right) \]

There are two main problems with the scaled adjacency-weight matrix is in this form. 
First, this matrix product is the sum of all neighboring nodes without inclusion of the source node (unless the graph contains self loops), so we can fix this by adding the identity matrix $I$.
Second,  $A$ is typically un-normalized and multiplication will change the scale. 
Therefore, we should row-wise normalize $A$ using the diagonal node degree matrix, $D$.
\cite{Kipf2017} proposes using a symmetric normalization to avoid these problems, reaching the following:

\[ f(L^{(n)},A) = \sigma \left( 
    \hat{D}^{-1/2}\hat{A}\hat{D}^{-1/2} L^{(n)}W^{(n)} \right) \]

To ensure that we capture the relation of node's own features, $A$ is replaced with $\hat{A} = A + I$, where $I$ is the identity matrix and $\hat{D}$ is the diagonal node matrix of $\hat{A}$.  

A graph convolution integrates data from the immediate neighborhood of a node. 
However, this graph is not encoded to be directional (the adjacency matrix is symmetrical).
Therefore, in our specific case, if there is more than one graph convolution, the network would effectively be allowed to look in to the future. 
A "fair" prediction for a one convolution GCN would actually be predicting two games ahead, and for a one convolution network three games ahead, etc.
Therefore, in our experiments we were careful to properly attach the label to each feature set from future game results.
Figure \ref{fig:gconv} explains how this model architecture impacts label choice.

\subsection{Chebyshev Polynomials}

Graph convolutions tend to fall into two categories: spectral and spatial \cite{Wu2020}.
Spectral approaches are derived from graph signal processing \cite{Shuman2013} and involve manipulating the scaled adjacency matrix of varying graph neighborhood sizes.  
At a high level, Chebyshev polynomials are an alternate way to perform graph convolutions using a spatial approach instead of the traditional spectral convolutions \cite{Velickovic2018}.
In this analysis we use a more traditional GCN model using spectral convolutions (\emph{gcn}) and a GCN model using Chebyshev polynomial filters to create convolutions (\emph{gcn-cheby})
 
\begin{figure}[t]
\centering
\includegraphics[width=0.90\linewidth]{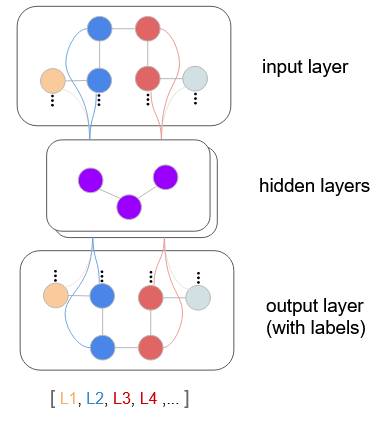}
\caption{
    GCN-WP network structure. The input layer (top) is formed from the graph produced by Algorithm \ref{alg:graph}. The convolutions are learned by the graph (middle), then the final layer is the same dimension as the original graph (bottom). L1 translates to label 1, etc. These are the predictions made by the model (win = 1, loss = 0).  }
\label{fig:graph}
\end{figure}

\section{Baseline Experiments}

\subsection{Machine Learning - Random Forest}

Random forest models are one of the most common and robust methods for classification in machine learning \cite{rf} and there are many open source implementations of the model \cite{pedregosa2011}. 
A random forest model trains $n$ decision trees and uses all of their classifications to vote on the final class of the answer, actually returning a class probability as number of classifiers predicting positive label over the total number of classifiers. 
They are more reliant on feature engineering than deep learning, however; but should still provide an appropriate baseline for the win prediction task \cite{Heaton2016}.

To be a fair classification challenge, to predict the label (win = 1, loss = 0) at $t_0$, we use data from the previous game ($t_{-1}$). 
Given that many win prediction models use skill updates based on past performance, we thought it was a valid assumption to use data from the last $n$ games a team plays.
Therefore instead of observing only the past game (at $t_{-1}$) and predicting the next, the features were collected over the past 1, 3 and 5 games. 
We set a constant, linear combination of features for each of the number of past game data used. 
In table \ref{tab:rfacc}, we show the results of the longest lookback, 5 games, on the \emph{delta} feature set described above has the highest performance.
This suggests that using past games and a relative team performance improve model accuracy.

\begin{table}[t]
    \begin{centering}
    \begin{tabular}{|c|c|c|c|} \hline
        Lookback & 1 game & 3 games & 5 games \\ \hline
        Raw & $0.563 \pm 0.009$ & $0.573 \pm 0.010$  &  $0.572 \pm 0.005$\\ \hline
        Delta & $0.560 \pm 0.009$ & $0.569 \pm 0.008$ & $\mathbf{0.578 \pm 0.012}$ \\ \hline
    \end{tabular}\\[1em]
    \caption{Win Prediction accuracy of Random Forest model}
    \label{tab:rfacc}
    \end{centering}
\end{table}

\subsection{Skill Modeling - SCOPE}

\begin{table}[t]
    \centering
    \begin{tabular}{|l|cccccc|} \hline 
        Base $K$ & 5 & 10 & 20 & 30 & 40 & 50  \\ \hline
        Cutoff & 1600 &  1650 &  1700 & 1750 & {} & {} \\ \hline
        Reduction & 0.1 & 0.2 & 0.3 & 0.4 & 0.5 & {} \\ \hline
        MoV func & none &  lin &  exp &  log & {} & {} \\ \hline
        $w90$ & 100 & 200 & 300 & 400 & 500 & {} \\ \hline
        Regression & 0 & 0.1 & 0.2 & 0.3 & 0.4 & {} \\ \hline
    \end{tabular} 
    \newline
    \vspace{1em}
    \newline
    \begin{tabular}{|c|c|c|} \hline
    \textbf{League} & LCS & LPL  \\ \hline
    Regress  & 0.4 & 0.4 \\ \hline
    Base $K$  & 40 & 40  \\ \hline
    Cutoff  & 1600 & 1700  \\ \hline
    Reduction  & 0.1 & 0.5 \\ \hline
    MoV  & none & lin  \\ \hline
    $w90$  & 100 & 500  \\ \hhline{|=|=|=|}
    \textbf{Test Acc} & 0.591 & 0.597  \\ \hline
    \end{tabular} \\[1em]
    \caption{On top, values tested for for SCOPE on LoL, below are the best values for respective leagues, the LCS and LPL.}
    \label{tab:scope_params}
\end{table}

SCOPE is a recently developed skill modeling framework that shows promise for win prediction \cite{scope}.
The authors show this method for it's efficacy on a Call of Duty League dataset, so we used the procedures they outlined to build a win prediction model for LoL.
Although we have data ranging from 2014, we decided to use data from the 3 most recent seasons to emulate the process that they used in \cite{scope}.
So, the 2018 season was used for initialization of the Elo scores, 2019 was used to find the best parameters for the model (validation) and the 2020 was the test set. 
We preformed the analysis on the regular season games of the LCS and the LPL.

There were some specifications of the model that were adapted to LoL, in particular Margin of Victory (MoV).
In LoL there is only one game mode played on one map, unlike in the CDL where teams compete across 3 different game modes in a best of 5. 
In each LoL game we used the difference in the wining team's kills vs. the losing team's kills. 
Another metric we debated using was the total gold difference between the teams. 
Kills is a popular metric used by broadcast analysts so we decided to use the kill difference as the MoV. 
Similarly to their procedure, we fit the MoV range to linear, square root, log and exponential functions. 
, the best of which are presented in 
For the validation run on the LPL there were 19 teams in 2019, and only 14 teams in 2018.
Instead of assuming these teams were below average skill level, we added them in at average ($1500$).

We used cross-validation to test a wide array of values, similar to process used in \cite{scope}, the results of which are presented in Table \ref{tab:scope_params}.
SCOPE selected similar parameters for both leagues, which is expected. 
Higher ranked teams in the LPL were in general more dominating which we can see by the score cutoff at $1700$ and $K$ reduction of $0.5$.
This means that teams above $1700$ had their $K$ update value cut in half, which made them much less susceptible to changes in score. 
In general the top ranked LPL teams are some of the best in the world and that league is also the largest of any region in 2020. 
This is an improvement on the random forest algorithm, and although does not have as high an accuracy as compared the original paper, we believe this was an effective application of the method. 

\begin{table}[b]
    \centering
    \begin{tabular}{|l|ccc|} \hline 
        Hidden Layer 1 & 32 & 64 & 128  \\ \hline
        Hidden Layer 2 & 32 & 64 & 128  \\ \hline
        Dropout & 0.1 & 0.25 & 0.5 \\ \hline
        Model & gcn & gcn\_cheby & {} \\ \hline
        Dataset & raw & delta & {} \\ \hline
    \end{tabular} \\[1em]
    \caption{Cross-validation parameters for the GCN-WP model}
    \label{tab:gcn-cv}
\end{table}

\section{Model Evaluation}

\subsection{GCN-WP Training and Testing}

The Chinese league, LPL, had double the games compared to any other league, so this league was fixed as a training network. 
The validation network is the LCK in Korea, which has the second most games.
Finally, the test network is the LCS, or the American league.
In order to avoid contaminating our model by switching the test, validation and training set, we decided to fix these leagues ahead of time. 
That way, we would avoid selecting which set would avoid the optimal accuracy because we likely would not have that choice in practice.

In addition we fixed this dataset to one input year. 
Although players and coaches may change slightly within a season, they could change significantly from season to season. 
Since the esports markets change very quickly, some teams are present one year and gone the next, this would leave sub-optimal network structure and prevent learning on the graph. 
Therefore, we decided to limit the dataset to the final year of data: 2020.

Using semi-supervised learning is beneficial because we can still use data at the beginning and end of the window which don't have results yet.
This is one key facet of the semi-supervised approach taken here that the other fully supervised models described above miss out on. 
As mentioned above, an assumption of Kipf's GCN is that the network is homogeneous. 
Given this modeling choice there is no difference between self past game nodes and opponent nodes.

The code for this model is an open source fork of Thomas Kipf's original GCN model \cite{gcn-github}, available here \cite{ajb-gcn-github}, implemented in python using Tensorflow \cite{tf2015}.
Algorithm \ref{alg:graph} was used to construct the adjacency matrix that fed in to a graph convolutional network. 
We tested various aspects of the model using a cross-validation set with 1 and 2 hidden layers. 
In addition, we varied the dropout of the network. 
Dropout is a reliable way to prevent overfitting in neural networks \cite{srivastava2014dropout}.
Grid search cross-validation were performed to determine the optimal features and model parameters.
The hyperparameters tested during cross-validation are listed in Table \ref{tab:gcn-cv}.

\begin{table}[t] 
\centering
\begin{tabular}{|c|c|} \hline
    \textbf{Model} & \textbf{Accuracy} \\ \hline 
    GCN-cheby (1 layer) + raw  & $0.541$  \\ \hline
    GCN  (1 layer) + delta & $0.551$  \\ \hline
    GCN-cheby (2 layer) + delta & $0.568$ \\ \hline
    Random Forest (lookback=5) + delta  & $0.578$  \\ \hline
    SCOPE (Elo) & $0.597$  \\ \hline
    GCN-cheby (1 layer) + delta & $\mathbf{0.619}$ \\ \hline
\end{tabular} \\[1em]
\caption{Best parameters from all models tested.}
\label{tab:gcn}
\end{table}

\subsection{Model Results and Comparison}

Table \ref{tab:gcn} presents the win prediction accuracy of all of the models tested. 
In general SCOPE and Random Forest performed well, but were outperformed by one construction of GCN-WP. 
In the following section we will analyze the various GCN-WP models and consider why  GCN-cheby (1 layer) + delta performed the best.

Chebyshev polynomials, although increasing compute time, increased model accuracy.
This was an interesting result especially since in the original GCN paper this method actually performed slightly worse \cite{Kipf2017}. 
Ensuring they were set to $1$ degree meant that the network shouldn't be integrating future game data since the degree is proportional to the neighborhood size of the convolution.
Given the feature representation of this data is more continuous, as opposed to word vector embeddings used in the paper, using these spacial convolutions likely has a more meaningful impact on the analysis of a node's neighbors.
For example, the actual distance of a feature node's total gold earned and it's neighbor's earned gold may have a direct numerical relationship. 
The winning team will likely have a much higher total gold and the spectral convolution with the Chebyshev polynomial should capture that. 

Using the delta dataset significantly improved model accuracy as well. 
Given that the graph convolutions are homogeneous, it is better that the features are relative between the winning and losing opponent. 
This is actually very similar to how the random forest model is encoded. 
A one layer GCN model will look at a neighborhood of $1$, so the node immediately before and after the labeled node. 
This way we are actually integrating data from $3$ games -- see the first row of figure 
This way, the model can determine the relative performance of nearby teams in relation to the class of the node that is being predicted. 

More layers reduced prediction accuracy. 
This is likely due to the fact that expanding the node neighborhood may not actually be beneficial in this case. 
A $K=2$ neighborhood would include teams unrelated to the team that is being predicted, so this would add noise the predictions for that particular node.
On the bottom row of Figure \ref{fig:gconv}, the graph convolution shown does not reveal the full neighborhood. 
It will also include opponents of opponents. 
The model then integrates that data in a convolution. 
It appears that this data may be too far away or noisy to correctly influence a prediction for the particular team-game node being classified. 

\section{Conclusions and Future Work}

In GCN-WP, a single team is not modeled over time, but instead this model learns a representation of the league structure and use that representation to predict success of teams in other leagues. 
This is both a novel and effective technique that allows for the maximum amount of data to be integrated for a prediction using semi-supervised learning.
This model has comparable, if not better, accuracy to other state of the art win prediction models. 
Although this use case is particularly well suited to LoL, this has applicability to other esports, like Overwatch, Counter Strike: Global Offensive, or even sports like Soccer or Basketball which have many leagues in addtion to the NBA.

In this case we did not assign a particular skill value to an individual team, like Elo or TrueSkill.
This model does not follow a particular team over time, but future work could use a calculated skill rating as a feature to boost the model's performance, similar to \cite{ghost-recon}.
In addition, we plan to expand the expressiveness of this model by using a heterogeneous (and directed) GCN model to represent the data \cite{Yang2020}.
The idea is that if each team node and enemy node is represented differently, the network will be able to differentiate which data belongs to an enemy and which data belongs to a particular team. 
This will enable us to use more convolutional layers looking backwards without the risk of leakage.

\bibliographystyle{IEEEtran}
\bibliography{main.bib}

\end{document}